\documentclass[10pt,twocolumn,letterpaper]{article}

\usepackage{iccv}
\usepackage{times}
\usepackage{epsfig}
\usepackage{graphicx}
\usepackage{amsmath}
\usepackage{amssymb}
\usepackage{subfigure}
\usepackage{multirow}
\usepackage{enumitem}
\usepackage{url}

\usepackage[pagebackref=true,breaklinks=true,letterpaper=true,colorlinks,bookmarks=false]{hyperref}

\newcommand{\minisection}[1]{\vspace{0.04in} \noindent {\bf #1}\ \ }

\iccvfinalcopy 

\def\iccvPaperID{340} 

\ificcvfinal\fi
\begin{document}

\title{RankIQA: Learning from Rankings for No-reference Image Quality Assessment}

\author{Xialei Liu\\
Computer Vision Center \\
Barcelona, Spain\\
{\tt\small xialei@cvc.uab.es}
\and
Joost van de Weijer\\
Computer Vision Center\\
Barcelona, Spain\\
{\tt\small joost@cvc.uab.es}
\and
Andrew D. Bagdanov\\
MICC, University of Florence \\
Florence, Italy\\
{\tt\small andrew.bagdanov@unifi.it}
}

\maketitle

\begin{abstract}
 We propose a no-reference image quality assessment
  (NR-IQA) approach that learns from rankings 
  (\emph{RankIQA}). To address the problem of limited IQA dataset size, we
  train a Siamese Network to rank images in terms of image quality by
  using synthetically generated distortions for which relative image
  quality is known. These ranked image sets can be automatically
  generated without laborious human labeling. We then use
  fine-tuning to transfer the knowledge represented in the trained
  Siamese Network to a traditional CNN that estimates absolute image
  quality from single images. We demonstrate how our approach can be
  made significantly more efficient than traditional Siamese Networks
  by forward propagating a batch of images through a single network
  and backpropagating gradients derived from all pairs of images in
  the batch. Experiments on the TID2013 benchmark show that we improve the state-of-the-art by over 5\%. Furthermore, on the LIVE benchmark we show that our approach is superior to existing NR-IQA techniques and that we even outperform the state-of-the-art in full-reference IQA (FR-IQA) methods without having to resort to high-quality reference images to infer IQA.
\end{abstract}

\section{Introduction}

Images are everywhere in our life. Unfortunately, they are often
distorted by the processes of acquisition, transmission, storage, and
external conditions like camera motion. Image Quality
Assessment (IQA)~\cite{wang2002image} is a technique developed to
automatically predict the perceptual quality of images. IQA estimates
should be highly correlated with quality assessments made by a range
of very many human evaluators (commonly referred to as the Mean
Opinion Score
(MOS)~\cite{sheikh2006statistical,ponomarenko2013color}). IQA has
been widely applied to problems where image quality is essential, like image
 restoration~\cite{katsaggelos2012digital}, image
super-resolution~\cite{van2006image}, and image
retrieval~\cite{yan2014learning}.

\begin{figure}
\centering
\includegraphics[width=8.0cm]{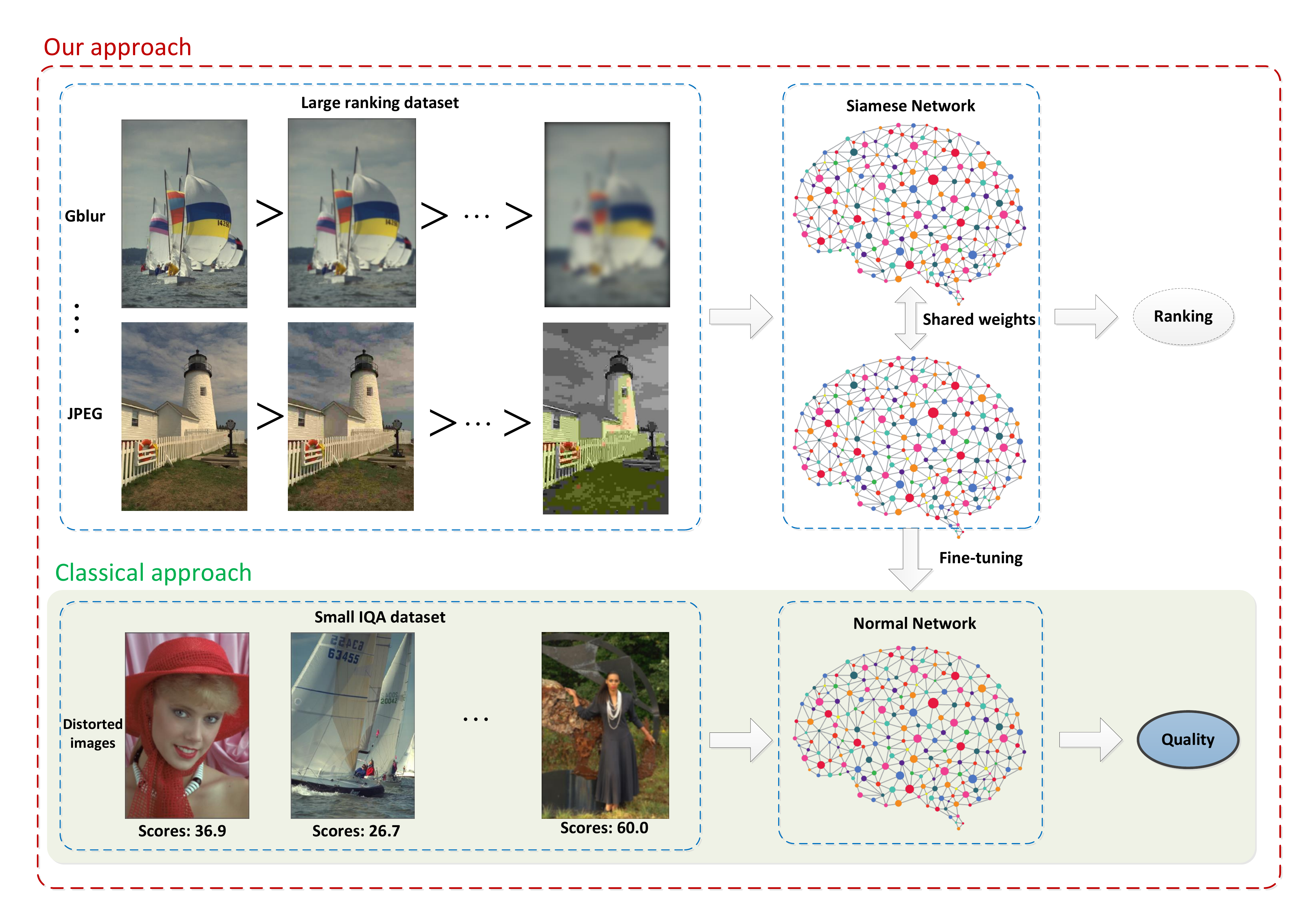}
\caption{The classical approach trains a deep CNN regressor
   \emph{directly} on the ground-truth. Our approach trains a network from an image
   \emph{ranking} dataset. These ranked images can be easily generated
   by applying distortions of varying intensities. The network parameters are then transferred to the
   regression network for fine-tuning. This allows for the training of deeper and wider networks.}
\label{fig_fw}
\end{figure}

IQA approaches are generally divided into three categories based on
whether the undistorted image (called reference image) or information about it is available:
full-reference IQA (FR-IQA), reduced-reference IQA (RR-IQA), and
no-reference IQA (NR-IQA). Research has mostly focussed on the more realist scenario of NR-IQA where the image
quality of an image without any reference image has to be estimated.
In NR-IQA, many methods focus on a specific
distortion~\cite{yan2013no,golestaneh2014no}, which limits the
applicability of these methods. Other methods consider a range
of distortions~\cite{moorthy2010two,saad2012blind,liu2014no,mittal2012no}.


Convolutional Neural Networks (CNNs) are having an enormous impact on
computer vision research and practice. Though they have been around
for decades~\cite{lecun1998gradient}, it wasn't until 2012, when
Krizhevsky et al.~\cite{krizhevsky2012imagenet} achieved spectacular
results with a CNN in the ImageNet competition, that they achieved
wide attention and adoption in the broader computer vision
community. The architectures of networks are getting deeper and deeper
with respect to the original AlexNet, with ResNet being an example of
very deep network architecture~\cite{he2015deep}. The result of this
trend is that state-of-the-art CNNs like AlexNet and ResNet have
hundred of millions of parameters and require massive amounts of data
to train from scratch (without overfitting).


The success of CNNs encouraged research exploring their potential
application to the NR-IQA problem. This research resulted in significant
improvements compared to previous hand-crafted approaches ~\cite{liang2016image,kang2014convolutional,kang2015simultaneous}.
The main problems these papers had to address is the absence of large datasets for IQA .
Especially as networks grow deeper and wider, the number of parameters increases dramatically. As a
consequence, larger and larger annotated datasets are required for
training. However, the annotation process for IQA image datasets
requires multiple human annotations for every image, and thus the
collection process is extremely labor-intensive and costly. As a
results, most available IQA datasets are too small
to be effective for training CNNs.

We propose an approach to address the absence of large datasets.
The main idea (see Fig.~\ref{fig_fw}) is that while human-annotated IQA data is difficult to
obtain, it is easy to \emph{generate} images that are
\emph{ranked} according to their image quality. That is, we can
generate image sets in which, though we do not have an absolute
quality measure for each generated image, for any \emph{pair} of
images we know which is of higher quality. We call this
\emph{learning from rankings} approach RankIQA, and with it we learn to rank image in terms of quality using Siamese Networks, and then we transfer knowledge
learned from ranked images to a traditional CNN fine-tuned on IQA data in order to improve the accuracy of IQA. The idea to learn IQA features from distorted reference images was proposed by Zhang et al. in a patent~\cite{zhang2015label}. In this paper we go beyond this patent in that we provide a detailed description of our method and experimentally verify the usefulness of pre-training networks using ranked datasets.

As a second contribution we
propose a method for efficient backpropagation in Siamese networks. The method
forwards a batch of images through a single network and then backpropagates gradients derived from
all pairs in the batch.
In extensive experiments on established IQA datasets
we show that learning from rankings significantly improves results, and that our efficient backpropagation
algorithm allows to train these networks better and faster than other training protocols, like hard-negative mining. The supplementary material and project page are available at \href{url}{https://github.com/xialeiliu/RankIQA}.

\section{Related work}
We briefly review the literature related to our approach. We focus on
distortion-generic NR-IQA since it is more generally applicable than
the other IQA research lines.

\minisection{Traditional NR-IQA approaches.}  Most traditional NR-IQA
can be classified into Natural Scene Statistics (NSS) methods and
learning-based methods. In NSS methods, the assumption is that images
of different quality vary in the statistics of responses to specific
filters. Wavelets~\cite{moorthy2010two}, DCT~\cite{saad2012blind} and
Curvelets~\cite{liu2014no} are commonly used to extract the features
in different sub-bands. These feature distributions are parametrized,
for example with the Generalized Gaussian
Distribution~\cite{sharifi1995estimation}. The aim of these methods is
to estimate the distributional parameters, from which a quality
assessment can be inferred. The authors of~\cite{mittal2012no} propose to extract NSS features in the spatial domain to obtain
significant speed-ups. In learning-based methods, local features are extracted and mapped to
the MOS using, for example, Support Machine Regression
or Neural Networks~\cite{chetouani2010novel}. The codebook
method~\cite{ye2012unsupervised,ye2012no} combines different features
instead of using local features directly. Datasets without MOS can be
exploited to construct the codebook by means of unsupervised learning,
which is particularly important due to of the small size of existing
datasets. Saliency maps~\cite{zhang2015som} can be used to model human
vision system and improve precision in these methods.

\minisection{Deep learning for NR-IQA.}  In recent years several works
have used deep learning for
NR-IQA~\cite{bianco2016use,kang2014convolutional,kang2015simultaneous}.
One of the main drawbacks of deep networks is the need for large
labeled datasets, which are currently not available for NR-IQA
research. To address this problem Kang et
al.~\cite{kang2014convolutional} consider small $32 \times 32$ patches
rather than images, thereby greatly augmenting the number of training
examples. The authors of~\cite{bosse2016deep,kang2015simultaneous}
follow the same pipeline. In~\cite{kang2015simultaneous} the authors
design a multi-task CNN to learn the type of distortions and image
quality simultaneously. Bianco at al.~\cite{bianco2016use} propose to
use a pre-trained network to mitigate the lack of training data. They
extract features from a pre-trained model fine-tuned on an IQA
dataset. These features are then used to train an SVR model to map
features to IQA scores. 

In our paper, we propose a radically different
approach to address the lack of training data: we use a large number
of automatically generated rankings of image quality to train a deep
network. This allows us to train much deeper and wider networks than
other methods in NR-IQA which train directly on absolute IQA data.


\minisection{Learning to rank.} These approaches learn a ranking
function from ground-truth rankings by minimizing a ranking
loss~\cite{chen2009ranking}. This function can then be applied to rank
test objects. The authors of~\cite{sculley2009large} adapt the
Stochastic Gradient Descent method to perform pairwise learning
to rank. This has been successfully applied to large
datasets. Combining ideas from ranking and CNNs, the Siamese network
architecture achieves great success on the face verification
problem~\cite{chopra2005learning}, and in comparing image
patches~\cite{zagoruyko2015learning}. The only other work which
applies rankings in the context of NR-IQA is~\cite{gao2015learning} in which they combine different hand-crafted
features to represent image pairs from the IQA dataset.

Our approach is different in that primarily we are not aiming to learn
rankings. Instead we use learning from rankings as a data augmentation
technique: we use easily obtainable datasets of ranked images to
train a large network, which is then fine-tuned for the task of
NR-IQA.


\minisection{Hard-negative mining for Siamese network training.} It is
well known that a naive approach to sampling pairs to training Siamese
networks is suboptimal. To address this problem several approaches to
hard-negative mining have been
proposed. In~\cite{simo2015discriminative}, they propose a hard
positive and hard-negative mining strategy to forward-propagate a set
of pairs and sample the highest loss pairs with back-propagation.
However, hard mining comes with a high computational cost (they report
an increase of up to 80\% of total computation
cost). In~\cite{schroff2015facenet} they propose semi-hard pair
selection, arguing that selecting hardest pairs can lead to bad local
minima. The batch size used is around 1800 examples, which again leads
to a considerable increase in computational
cost. In~\cite{wang2015unsupervised} the authors take a batch of
pairs as input and choose the four hardest negative samples within the
mini-batch. To solve for a bad local optimum, \cite{song2015deep}
optimize a smooth upper bound loss function to take advantage of all
possible pairs in the mini-batch.

In contrast with these works, we propose a method for efficient
Siamese backpropagation which does not depend on hard-negative
selection. Instead, it considers all possible pairs in the
mini-batch. This has the advantage that the main computational
bottleneck in training deep networks, namely the forward-propagation
of images through the network, is optimally exploited.

\section{Learning from rankings for NR-IQA}
\label{sec:ranking}

In this section we describe our approach to exploiting synthetically
generated rankings for NR-IQA. We first lay out a general framework
for our approach, then describe how we use a Siamese network
architecture to learn from rankings. Finally, in
section~\ref{sec:efficient-siamese} we show how backpropagation for
training Siamese networks from ranked samples can be made
significantly more efficient.

\subsection{Overview of our approach}
The lack of large IQA datasets motivates us to propose a new strategy
to take advantage of large, \emph{unlabelled} databases from which we
can generate images ranked by image quality. Our approach is based on
the observation that, given a set of arbitrary reference images, it is
very easy to apply image distortions to generate a \emph{ranking}
image dataset. As an example, given a reference image we can apply
various levels of Gaussian blur. The set of images which is thus
generated can be easily ranked because we do know that adding Gaussian
blur (or any other distortion) always deteriorates the quality
score. Note that in such set of ranked images we do not have any
absolute IQA scores for any images -- but we do know for any pair of
images \emph{which is of higher quality}.

After learning on these ranked images, we can use fine-tuning on small
image quality datasets in order to address the IQA problem. The
difference between our approach and the straightforward, classical
approach~\cite{kang2015simultaneous} is shown in
Fig.~\ref{fig_fw}. The standard approach trains a shallow network
directly on the IQA dataset to estimate IQA score from images. Due to
the limited data only few layers can be used, which limits
accuracy. Since we have access to much larger datasets with ranked
images, we can now train deeper and wider networks to learn a distance
embedding. Next we follow this by fine-tuning for domain adaptation to
the absolute IQA problem. The overall pipeline of our approach is:
\begin{enumerate}[leftmargin=*]
\setlength\itemsep{0em}
\item \textbf{Synthesize ranked images.} Using an arbitrary set of
  images, we synthetically generate deformations of these
  images over a range of distortion intensities. The absolute
  distortion amount for each image is not used in subsequent steps,
  but within each deformation type we know, for any pair of images,
  which is of higher quality. See section~\ref{sec:datasets} for a
  description of the datasets used for generating ranked images and
  the distortions applied.
\item \textbf{Train Siamese network for ranking.} Using the set of
  ranked images, we train a Siamese network described in the next
  section using the efficient Siamese backpropagation technique
  proposed in section~\ref{sec:efficient-siamese}. The result is a Siamese
  network that \emph{ranks} images by image quality. 
\item \textbf{Fine-tune on IQA data.} Finally, we extract a single
  branch of the Siamese network (we are interested at this point in
  the representation learned in the network, and not in the ranking
  itself), and fine-tune it on available IQA data. This effectively
  calibrates the network to output IQA measurements.
\end{enumerate}

\subsection{Siamese networks for ranking}
Here we introduce the Siamese network~\cite{chopra2005learning} to learn from
image rankings, which is a
network with two identical network branches and a loss module. The two
branches share weights during training. Pairs of images and labels are
the input of the network, yielding two outputs which are passed to the
loss module. The gradients of the loss function with respect to all
model parameters are computed by backpropagation and updated with the
stochastic gradient method.



Specifically, given an image $x$ as the input of the network, the
output feature representation of $x$, denoted by $f(x;\theta)$, is
obtained by capturing the activation in the last layer. Here $\theta$
are the network parameters, and we will use $y$ to denote the ground truth value for the image which for image quality assessment is its quality score. Consequently, in our Siamese networks the output of the
final layer is always a single scalar which we want to be indicative
of the image quality.  Since our aim is to rank the images, we apply
the pairwise ranking hinge loss:
\begin{equation} \label{eq:ranking}
L({x}_{1},{ x }_{ 2 };\theta) = \max{(0,f(x_{2};\theta)-f( x_{ 1 };\theta)+\varepsilon )}
\end{equation}
where $\varepsilon$ is the margin. Here we assume without loss of generality that the rank of ${x}_{1}$
is higher than ${x}_{2}$. The gradient of the loss in
Eq.~\ref{eq:ranking} is given by:
\begin{equation}
\resizebox{0.8\columnwidth}{!}
{$
{ \nabla  }_{ \theta  }L =
\begin{cases}
  0 ~~~~~~~~~~~~~~~~~  \text{if } f\left( { x }_{ 2 };\theta \right) - f\left( { x
    }_{ 1 };\theta  \right) + \varepsilon \le 0 \\
  { \nabla }_{ \theta }f\left( { x }_{ 2 };\theta \right) -{ \nabla
  }_{ \theta }f\left( { x }_{ 1 };\theta \right) ~~~ \text{otherwise}.
\end{cases}
$}
\end{equation}
In other words, when the outcome of the network is in accordance with
the ranking, the gradient is zero. When the outcome of the network is
not in accordance we decrease the gradient of the higher and add the
gradient of the lower score. Given this gradient of $L$ with respect
to model parameters $\theta$, we can train the Siamese network using
Stochastic Gradient Descent (SGD).

%

\subsection{Efficient Siamese backpropagation}
\label{sec:efficient-siamese}

One drawback of Siamese networks is the redundant
computation. Consider all possible image pairs constructed from three
images. In a standard implementation all three images are passed twice
through the network, because they each appear in two pairs. Since both
branches of the Siamese network are identical, we are essentially
doing twice the work necessary since any image need only be passed
\emph{once} through the network. It is exactly this idea that we
exploit to render backpropagation more efficient. In fact, nothing prevents us from considering all possible
pairs in the mini-batch, without hardly any additional computation.
We add a new layer to the network that generates all possible pairs in
a mini-batch at the end of the network right before computing the
loss. This eliminates the problem of pair selection and boosts
efficiency.



To approximate the speed-up of efficient Siamese backpropagation
consider the following. If we have one reference image distorted by
$n$ levels of distortions in our training set, then for a traditional
implementation of the Siamese network we would have to pass a total of
${ n }^{ 2 }-n$ images through the network -- which is twice the
number of pairs you can generate with $n$ images. Instead we propose
to pass all images only a single time and consider all possible pairs
only in the loss computation layer.  This reduces computation to just
$n$ passes through the network. Therefore, the speed-up is equal to:
$\frac { { n }^{ 2 }-n }{ n } = { n -1 }$. In the best scenario $n=M$,
where $M$ is equal to the number of images in the mini-batch, and
hence the speed-up of this method would be in the order of the
mini-batch size. Due to the high correlation among the set of all
pairs in a mini-batch, we expect the final speedup in convergence to
be lower.




To simplify notation in the following, we let
$\hat{y}_i = f(x_i;\theta)$, where $f(x_i;\theta)$ is the output of a
single branch of the Siamese network on input $x_i$. Now, for one pair
of inputs the loss is:
\begin{equation} \label{eq:general}
L(x_1,x_2,l_{12}; \theta) = g(\hat{y}_1,\hat{y}_2,l_{12}),
\end{equation}
where $l_{12}$ is a label indicating the relationship between image
$1$ and $2$ (for example, for ranking it indicates whether $x_1$ is of
higher rank than $x_2$), and
$\theta = \{\theta_1, \theta_2, \ldots, \theta_k\}$ are all model
parameters in the Siamese network. We omit the $\theta$ dependency in
$g$ for simplicity. The gradient of this loss function with respect to
the model parameter $\theta$ is:
\begin{equation}
  \nabla_{\theta} L=
  \frac{ \partial g(\hat{y}_1,\hat{y}_2,l_{12}) }
  {\partial \hat{y}_1 }
   \nabla_{\theta} \hat{y}_1
  +
  \frac{ \partial g(\hat{y}_1,\hat{y}_2,l_{12}) }
  {\partial \hat{y}_2 }
   \nabla_{\theta} \hat{y}_2.
\end{equation}
This gradient of $L$ above is a sum since the model parameters
  are shared between both branches of the Siamese network and
  $\hat{y}_1$ and $\hat{y}_2$ are computed using exactly the same
  parameters.

Considering all pairs in a mini-batch, the loss is:
\begin{equation}
L = \sum_{i=1}^{M}{ \sum _{ j > i }^{ M }{ g(\hat{y}_i,\hat{y}_j,l_{ij}) }}
\end{equation}
The gradient of the mini-batch loss with respect to parameter $\theta$
can then be written as:
\begin{equation}
\nabla_{\theta} L  =
\sum_{ i=1 }^{ M } \sum_{ j > i }^M
 \frac{ \partial g(\hat{y}_i,\hat{y}_j, l_{ij}) }
      { \partial \hat{y}_i}
      \nabla_{\theta} \hat{y}_i
 +
 \frac{ \partial g(\hat{y}_i,\hat{y}_j,l_{ij}) }
      { \partial \hat{y}_j }
      \nabla_{\theta} \hat{y}_j
\label{eq:loss2}
\end{equation}

%
%
We can now express the gradient of the loss function of the mini-batch in matrix form as:
\begin{equation}
\nabla _\theta  L = \left[ {\begin{array}{*{20}c}
   {\nabla _\theta  \hat y_1 } & {\nabla _\theta  \hat y_2 } &  \ldots  & {\nabla _\theta  \hat y_M }  \\
\end{array}} \right]P{\bf{1}}_M \label{eqn:minibatch-gradient}
\end{equation}
where $\mathbf{1}_M$ is the vector of all ones of length $M$. For a
standard single-branch network, we would average the gradients for all
batch samples to obtain the gradient of the mini-batch. This is
equivalent to setting $P$ to the identity matrix
in Eq.~\ref{eqn:minibatch-gradient} above. For Siamese networks where we
consider all pairs in the mini-batch we obtain Eq.~\ref{eq:loss2} by
setting $P$ to:
\begin{equation}
\arraycolsep=1.5pt
P = \left[
    \begin{matrix}
      0 & \frac { \partial g(\hat{y}_1,\hat{y}_2,l_{12}) }{ \partial \hat{y}_1 }  & \cdots  & \frac{\partial g(\hat{y}_1,\hat{y}_M,l_{1M})}{ \partial \hat{y}_1 }  \\
      \frac { \partial g(\hat{y}_1,\hat{y}_2,l_{12}) }{ \partial \hat{y}_2 } & 0 & \cdots & \frac { \partial g(\hat{y}_2,\hat{y}_M,l_{2M}) }{ \partial \hat{y}_2 } \\
      \vdots  & \vdots  & \ddots  & \vdots  \\
      \frac { \partial g(\hat{y}_1,\hat{y}_M,l_{1M}) }{ \partial \hat{y}_M } & \cdots &   \cdots & 0 \end{matrix} \right] \label{eq:P_matrix}
\end{equation}
The derivation until here is valid for any Siamese loss function of
the form Eq.~\ref{eq:general}. For the specific case of the ranking
hinge loss replace $g$ in Eq.~\ref{eq:P_matrix} with
Eq.~\ref{eq:ranking} and obtain:
\begin{equation}\label{eq:fast}
  P =  \left[ \begin{matrix} 0 & { a }_{ 12 } & \cdots  & { a }_{ 1M } \\ { a }_{ 21 } & 0 & \cdots  & { a }_{ 2M } \\ \vdots  & \vdots  & \ddots  & \vdots  \\ { a }_{ M1 } & \cdots  & { a }_{ M(M-1) } & 0 \end{matrix} \right],
\end{equation}
where
\begin{equation}\label{eqn:coefficients}
 a_{ij} =
\begin{cases}
0 & \text{if } l_{ij} \left( \hat{y}_j -  \hat{y}_i \right) + \varepsilon  \le  0   \\
l_{ij} & \text{otherwise}
\end{cases}
\end{equation}
and $l_{ij} \in \left\{ { - 1,0,1} \right\}$ where $1$ (-1) indicates
that ${y}_i > {y}_j$ (${y}_i < {y}_j$), and $0$ that ${y}_i = {y}_j$
or that they cannot be compared as is the case with images
corrupted with different distortions.


The above case considered a single distortion. Suppose instead we have
$D$ types of distortions in the training set. We can then compute the
gradient of the loss using a block diagonal matrix as:
\begin{equation}
\nabla _\theta  L = \left[ {\begin{array}{*{20}c}
   {\nabla _\theta  \hat Y^1 }    \ldots   {\nabla _\theta  \hat Y^D }  \\
\end{array}} \right]\left[ {\begin{array}{*{20}c}
   {A^1 } & 0 &  \cdots  & 0  \\
   0 & {A^2 } &  \cdots  & 0  \\
    \vdots  &  \vdots  &  \ddots  &  \vdots   \\
   0 & 0 &  \cdots  & {A^d }  \\
\end{array}} \right]{\bf{1}}_M 
\end{equation}
where
\begin{eqnarray}
\nonumber
{ A }^{ d } &=& \left[
\begin{matrix}
0 & { a }_{ 12 }^{ d } & \cdots  & { a }_{ 1M^d }^{ d } \\ { { a }_{ 21 }^{ d } } & 0 & \cdots  & { a }_{ 2M^d }^{ d } \\
\vdots  & \vdots  & \ddots  & \vdots  \\
{ { a }_{ M^d1 }^{ d } } & \cdots  & { a }_{ M^d(M^d-1) }^{ d } & 0
\end{matrix}
\right], \mbox{ and} \\
\nabla _\theta  \hat Y^d  &=& \left[
\begin{matrix}
{\nabla _\theta  \hat y_{d1} } &  \cdots  & {\nabla _\theta  \hat y_{dM} }  \\
\end{matrix}
\right].    
\end{eqnarray}
where ${\hat y }_{ mn }$ refers to the network output of the $n^{th}$ image with the $m^{th}$
distortion, $d \in \{1, 2, \ldots, D\}$, and $M = \sum {M^d }$ where
$M^d$ is the number of images with distortion $d$ in the mini-batch.
In the definition of $A^d$ above, the $a^d_{ij}$ are the gradient
coefficients as in
Eq.~\ref{eqn:coefficients}. 


\subsection{Fine-tuning for NR-IQA}
\label{sec:fine-tuning}

After training a Siamese network to rank distorted images, we then
extract a single branch from the network for fine-tuning. Given $M$
images in mini-batch with human IQA measurements, we denote the
ground truth quality score of the $i$-th image as $y_i$, and the
predicted score from the network is $\hat{y}_i$, as above. We fine-tune
the network using squared Euclidean distance as the loss
function in place of the ranking loss used for the Siamese network:
\begin{equation}
\label{eul}
L(y_i, \hat{y}_i) = \frac { 1 }{ M } \sum _{ i=1 }^{ M } (y_{i}-\hat{y}_{i})^2
\end{equation}

\section{Experimental results}
\label{sec:experiment}

In this section we report on a number of experiments designed to
evaluate the performance of our approach with respect to baselines and
the state-of-the-art in IQA. 

\subsection{Datasets}
\label{sec:datasets}
We use two types of datasets in our experiments: generic, non-IQA
datasets for generating ranked images to train the Siamese network,
and IQA datasets for fine-tuning and evaluation.

\minisection{IQA datasets.}  We perform experiments on two standard
IQA benchmark datasets.  The \textbf{LIVE}~\cite{live2} consists of
808 images generated from 29 original images by distorting them with
five types of distortion: Gaussian blur (GB), Gaussian noise (GN),
JPEG compression (JPEG), JPEG2000 compression (JP2K) and fast fading
(FF). The ground-truth Mean Opinion Score for each image is in the
range [0, 100] and is estimated using annotations by 161 human
annotators.The \textbf{TID2013}~\cite{ponomarenko2013color} dataset consists of
25 reference images with 3000 distorted images from 24 different
distortion types at 5 degradation levels. Mean Opinion Scores are in
the range [0, 9]. Distortion types include a range of noise,
compression, and transmission artifacts. See the original publication
for the list of specific distortion types.

\minisection{Datasets for generating ranked pairs.} To test on the
LIVE database, we generate four types of distortions which are widely
used and common : GB, GN, JPEG, and JP2K. To test on TID2013, we
generate 17 out of a total of 24 distortions (apart from \#3,
\#4,\#12, \#13, \#20, \#21, \#24). For the distortions which we could
not generate, we apply fine-tuning from the network trained from the
other distortions. This was found to yield satisfactory results.
We use two datasets for generating ranked image pairs.  

The
\textbf{Waterloo} dataset consists of 4,744 high quality natural
images carefully chosen from the Internet. Additionally, we use the
validation set of the \textbf{Places2}~\cite{places} dataset of 356
scene categories. There are 100 images per category in the validation
set, for a total 36500 images. After distortion, we have many more
distorted images for learning an image quality ranking embedding. The
aim of using this dataset is to demonstrate that high-quality ranking
embeddings can be learned using datasets not specifically designed for
the IQA problem.




\begin{figure}[tb]
  \centering
  \includegraphics[width = 7.0cm]{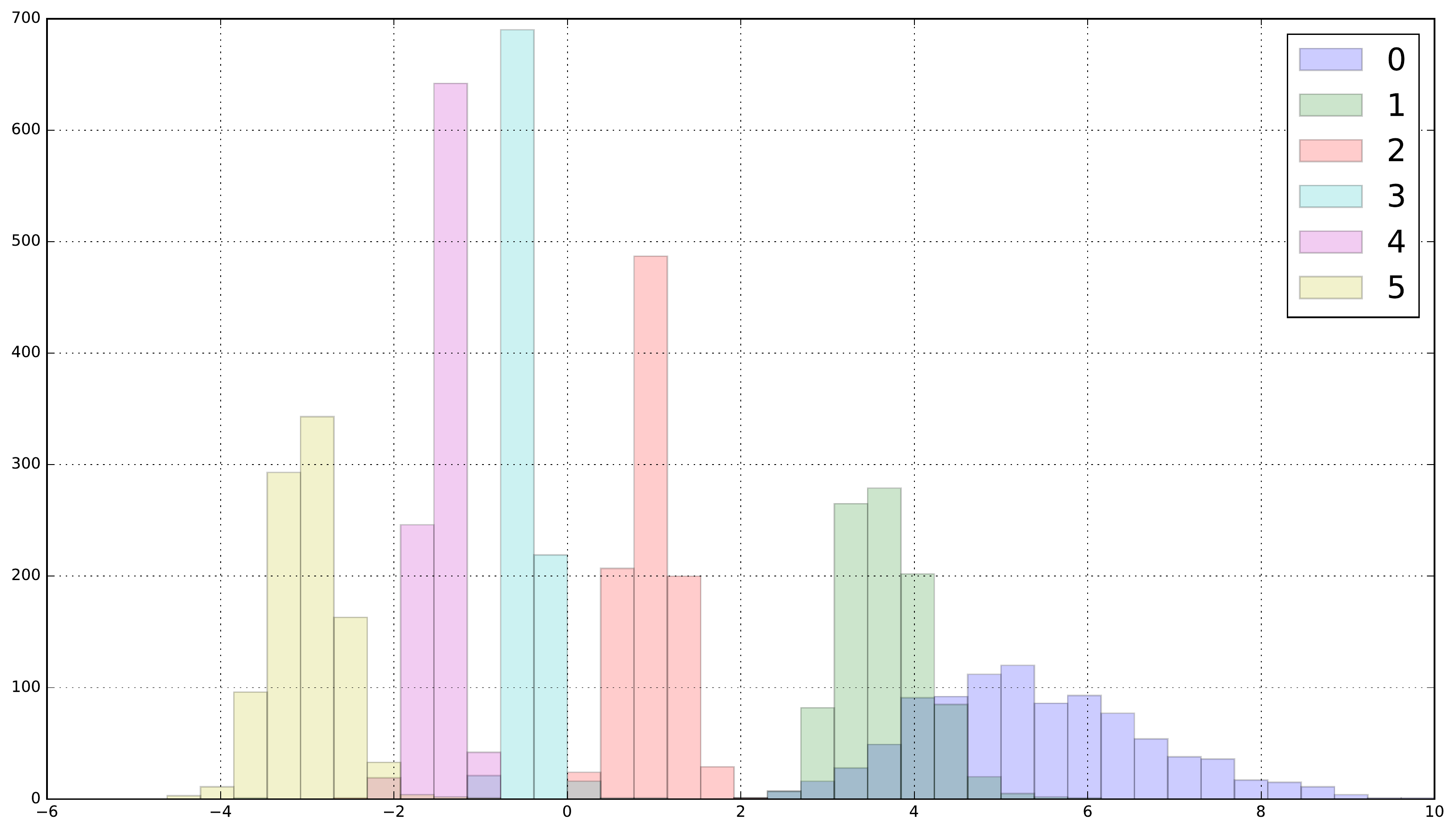}

  \caption{Siamese network output for JPEG  distortion considering 6 levels. This graphs illustrate the fact that the Siamese network
  successfully manages to separate the different distortion levels.}
  \label{fig:place}
\end{figure}

\subsection{Experimental protocols}

We investigate a number of network architectures and use standard IQA
metrics to evaluate performance.


\minisection{Network architectures.} We evaluate three typical network
architectures varying from shallow to deep. We refer to them as:
Shallow, AlexNet~\cite{krizhevsky2012imagenet}, and
VGG-16~\cite{simonyan2014very}. The shallow network has four convolutional layers and one fully connected layer.  For AlexNet
and VGG-16 we only change the number of
outputs since our objective is to assign one score for each
distorted image.


\minisection{Strategy for training and testing.} We randomly sample
sub-images from the original high resolution
images. We do this instead of scaling to avoid introducing
distortions caused by interpolation or filtering. The size of sampled
images is determined by each network. However, the
large size of the input images is important since input sub-images
should be at least 1/3 of the original images in order to capture
context information. This is a serious limitation of the patch
sampling approach that samples very small $32 \times 32$ patches from
the original images. In our experiments, we sample $227 \times 227$
and $224 \times 224$ pixel images, depending on the network. We use the Caffe~\cite{jia2014caffe} framework and train using mini-batch Stochastic Gradient Descent (SGD) with an initial learning rate of 1e-4 for efficient Siamese network training and 1e-6 for fine-tuning. Training rates are decreased by a factor of 0.1 every 10K iterations for a total of 50K iterations. For both training phases we use $\ell_2$ weight decay (weight 5e-4). During training we sample a single subimage from each training image per epoch. 

When testing, we randomly sample 30 sub-images from the original images as suggested in~\cite{bianco2016use}. The average of the outputs of the
sub-regions is the final score for each image.

\minisection{Evaluation protocols.} Two evaluation metrics are
traditionally used to evaluate the performance of IQA algorithms: the
Linear Correlation Coefficient (LCC) and the Spearman Rank Order
Correlation Coefficient (SROCC). LCC is a measure of the linear
correlation between the ground truth and the predicted quality
scores. Given $N$ distorted images, the ground truth of $i$-th image
is denoted by $y_i$, and the predicted score from the network is
$\hat{y}_i$. The LCC is computed as:
\begin{equation}
LCC = \frac{\sum_{ i=1 }^{ N }  ( { y }_{ i }-\overline { y } ) ( { \hat{y} }_{ i }-\overline { \hat{y} } )}
           {\sqrt{\sum _i^N ( { y }_{ i }-\overline { y } )^{ 2 } } \sqrt{ \sum _{ i }^{ N } ( \hat{ y }_{ i }-\overline { \hat{y} })^2  }}
\end{equation}
where  $\overline { y }$ and  $\overline { \hat{y} }$ are the means of the
ground truth and predicted quality scores, respectively.

Given $N$ distorted images, the SROCC is computed as:
\begin{equation}
  SROCC = 1 - \frac { 6\sum _{ i=1 }^{ N }{ { \left( { v }_{ i }-{ p }_{ i } \right)  }^{ 2 } }  }{ N\left( { N }^{ 2 }-1 \right)  },
\end{equation}
where $v_i$ is the \emph{rank} of the ground-truth IQA score $y_i$ in
the ground-truth scores, and $p_i$ is the \emph{rank} of $\hat{y}_i$
in the output scores for all $N$ images. The SROCC measures the
monotonic relationship between ground-truth and estimated IQA.



\begin{figure}[tbp]
\centering
\includegraphics[width = 6.0cm]{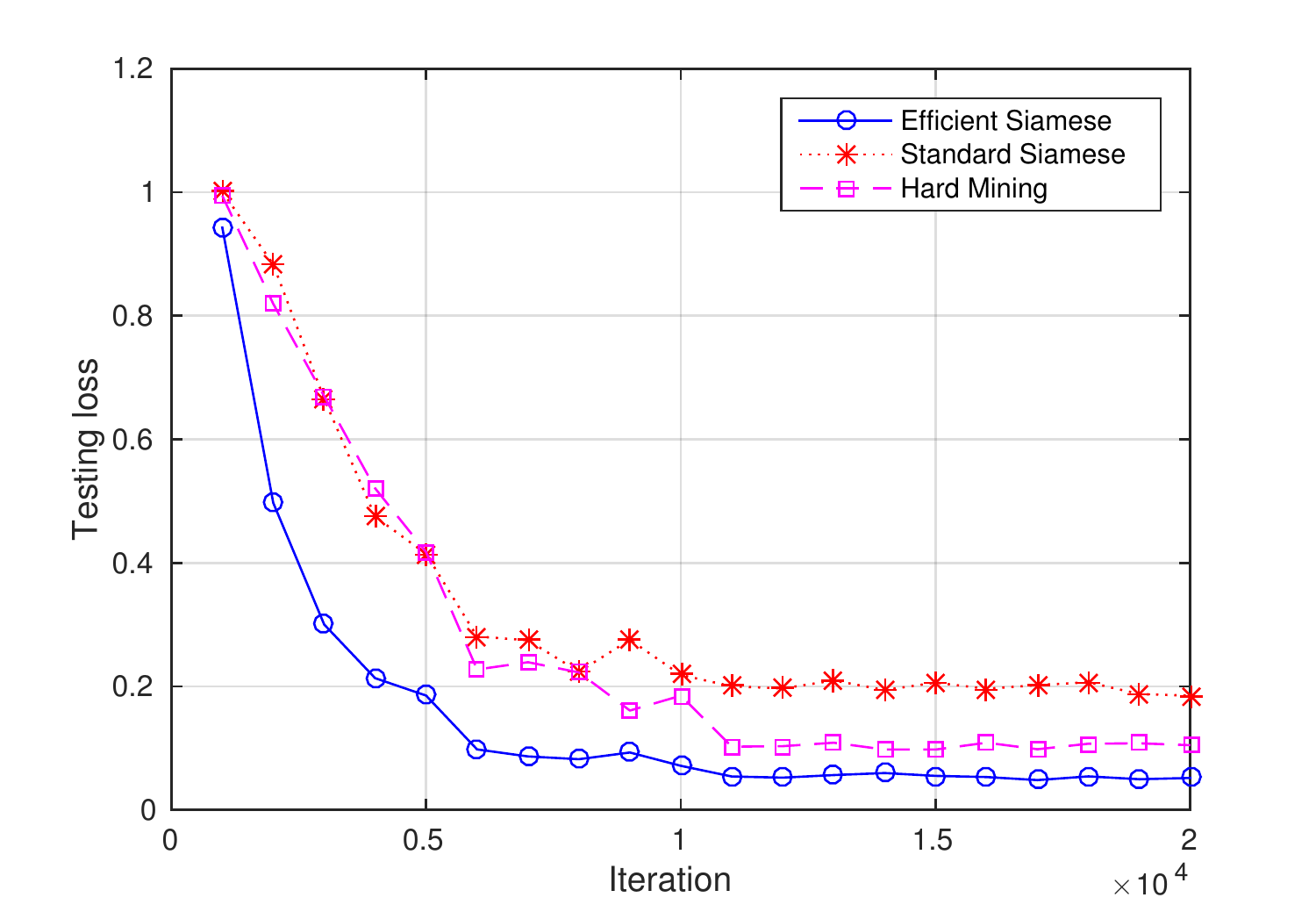}

\caption{Convergence of
  ranking loss on JPEG distortion for our approach versus standard
  Siamese and hard-negative mining. } 
  \label{fig:fast}
\end{figure}




\subsection{Learning NR-IQA from rankings}
We performed a number of experiments to evaluate the ability of
Siamese networks to learn to capture image distortions from a large
dataset of image quality rankings.  In addition, we measure the
impact of the efficient Siamese backpropagation approach described in
section~\ref{sec:efficient-siamese}.


\begin{table}[tbp]
\begin{center}
\resizebox{0.7\columnwidth}{!}{%
\begin{tabular}{c|ccc}
\hline
    & \textbf{Shallow} & \textbf{AlexNet} & \textbf{VGG-16} \\ \hline

\textbf{LCC}   & 0.930   & 0.945   & 0.973  \\ 
\textbf{SROCC}    & 0.937   & 0.949   & 0.974  \\ \hline
\end{tabular}}

\end{center}
\caption{SROCC and LCC for our approach on LIVE using different networks.}
\label{my-label}
\end{table}

\begin{table*}[tbp]
\begin{center}

\resizebox{1.9\columnwidth}{!}{%
\begin{tabular}{c|ccccccccccccc}
\hline
Method                               & \#01                               & \#02                               & \#03                               & \#04                               & \#05                               & \#06                               & \#07                               & \#08                               & \#09                               & \#10                               & \#11                                & \#12                      & \#13                               \\ \hline
BLIINDS-II~\cite{saad2012blind}                           & 0.714                              & 0.728                              & 0.825                              & 0.358                              & 0.852                              & 0.664                              & 0.780                              & 0.852                              & 0.754                              & 0.808                              & 0.862                               & 0.251                     & 0.755                              \\
BRISQUE~\cite{mittal2012no}                              & 0.630                              & 0.424                              & 0.727                              & 0.321                              & 0.775                              & 0.669                              & 0.592                              & 0.845                              & 0.553                              & 0.742                              & 0.799                               & 0.301                     & 0.672                              \\
CORNIA-10K ~\cite{ye2012unsupervised}                           & 0.341                              & -0.196                             & 0.689                              & 0.184                              & 0.607                              & -0.014                             & 0.673                              & \textbf{0.896}                              & 0.787                              & 0.875                              & 0.911                               & 0.310                     & 0.625                              \\
HOSA   ~\cite{xu2016blind}                           & 0.853                              & 0.625                              & 0.782                              & 0.368                              & \textbf{0.905}                              & 0.775                              & 0.810                              & 0.892                     & 0.870                              & 0.893                              & \textbf{0.932}                               & \textbf{0.747}            & 0.701                              \\ \hline
Baseline                      & 0.605                              & 0.468                              & 0.761                              & 0.232                              & 0.704                              & 0.590                              & 0.559                              & 0.782                              & 0.639                              & 0.772                              & 0.817                               & 0.571                     & 0.725                              \\
RankIQA    & \textbf{0.891} & \textbf{0.799} & \textbf{0.911} & \textbf{0.644} & 0.873 & \textbf{0.869} &\textbf{0.910} & 0.835          & \textbf{0.894} & \textbf{0.902} & 0.923  & 0.579 & 0.431          \\
RankIQA+FT     & 0.667          & 0.620          & 0.821          & 0.365          & 0.760          & 0.736          & 0.783          & 0.809          & 0.767          & 0.866          & 0.878           & 0.704 & \textbf{0.810} \\ \hline
Method                               & \#14                               & \#15                               & \#16                               & \#17                               & \#18                               & \#19                               & \#20                               & \#21                               & \#22                               & \#23                               & \multicolumn{1}{c|}{\#24}          & \multicolumn{2}{c}{ALL}                                        \\ \hline
BLIINDS-II     ~\cite{saad2012blind}                       & 0.081                              & 0.371                              & 0.159                              & -0.082                             & 0.109                              & 0.699                              & 0.222                              & 0.451                              & 0.815                              & 0.568                              & \multicolumn{1}{c|}{0.856}          & \multicolumn{2}{c}{0.550}                                      \\
BRISQUE   ~\cite{mittal2012no}                            & 0.175                              & 0.184                              & 0.155                              & 0.125                              & 0.032                              & 0.560                              & 0.282                              & 0.680                              & 0.804                              & 0.715                              & \multicolumn{1}{c|}{0.800}          & \multicolumn{2}{c}{0.562}                                      \\
CORNIA-10K    ~\cite{ye2012unsupervised}                       & 0.161                              & 0.096                              & 0.008                              & 0.423                              & -0.055                             & 0.259                              & 0.606                              & 0.555                              & 0.592                              & 0.759                              & \multicolumn{1}{c|}{0.903}          & \multicolumn{2}{c}{0.651}                                      \\
HOSA  ~\cite{xu2016blind}                                & 0.199                              & 0.327                              & 0.233                              & 0.294                              & 0.119                              & 0.782                              & 0.532                              & 0.835                              & \textbf{0.855}                     & \textbf{0.801}                     & \multicolumn{1}{c|}{\textbf{0.905}} & \multicolumn{2}{c}{0.728}                                      \\ \hline
Baseline & 0.187          & 0.308          & 0.062          & 0.546          & 0.383          & 0.500          & 0.420          & 0.700          & 0.611          & 0.573          & \multicolumn{1}{c|}{0.742}          & \multicolumn{2}{c}{0.612}                                      \\
RankIQA   & 0.463          & \textbf{0.693} & \textbf{0.321} & \textbf{0.657} & 0.622          & \textbf{0.845} & 0.609          & \textbf{0.891} & 0.788          & 0.727          & \multicolumn{1}{c|}{0.768}          & \multicolumn{2}{c}{0.623}                                      \\
RankIQA+FT                          & \textbf{0.512}                     & 0.622                              & 0.268                              & 0.613                              & \textbf{0.662}                     & 0.619                              & \textbf{0.644}                     & 0.800                              & 0.779                              & 0.629                              & \multicolumn{1}{c|}{0.859}          & \multicolumn{2}{c}{\textbf{0.780}}                             \\ \hline
\end{tabular} }

\end{center}
\caption{Performance evaluation (SROCC) on the entire TID2013 database. The baseline approach is VGG16 fine-tuned directly on TID2013 data without using the Siamese net, RankIQA is VGG16 only fine-tuned for ranking using Siamese net on generated ranking data, and RankIQA+FT is our learning-from-ranking approach further fine-tuned on TID2013 data.}
\label{ind_dis}
\end{table*}

\minisection{Siamese networks and IQA discrimination.}  To
demonstrate the ability of our ranking networks to discriminate image
quality, we trained our Siamese network on the Places2 validation set
(without applying fine-tuning on IQA data) corrupted with five levels of a
single distortion. We then used that network to predict image quality
for synthetically-distorted images from the Waterloo dataset corrupted
using the same five levels of the same distortion. The network outputs
are plotted as histograms in Fig.~\ref{fig:place} for the JPEG distortion\footnote{Graphs for the other distortions are in the supplementary material.}. In the plot, we
divide the observations according to the true distortion level
(indicated by the color of the histogram). The model discriminates different levels of distortions on
Waterloo, even though the acquisition process and the scenes of the two
datasets are totally different.

\minisection{Efficient Siamese backpropagation.}  The objective of
this experiment is to evaluate the efficiency of our Siamese
backpropagation method.  We compare our method to both standard random
pair sampling, and a hard-negative mining method similar
to~\cite{simo2015discriminative}.\footnote{We experimented with several
  hard-negative mining methods and found this to work best.}
   For standard random pair sampling we randomly choose 36
pairs for each mini-batch from the training sets. For the hard
negative mining strategy we start from 36 pairs in a mini-batch, and
gradually increase the number of hard pairs every 5000 iterations. For
our method we pass 72 images in each mini-batch. With these settings
the computational costs of all three methods is equal, since at each
iteration 72 images are passed through the network. We use AlexNet for
this experiment. The comparison of convergence rates on JPEG\footnote{Results for the other distortions are in the supplementary material.} is shown in Fig.~\ref{fig:fast}. The efficient Siamese backpropagation not only converges much faster, but also converges to a considerably lower
loss.

\minisection{Network performance analysis.} Here we evaluate whether we can learn a useful image representation from large image ranking datasets. We randomly split on the original, high-quality images before distortion from the LIVE dataset
into 80\% training and 20\% testing samples and compute the average
LCC and SROCC scores on the testing set after training to
convergence. This process is repeated ten times and the results are
averaged. In Table~\ref{my-label} we compare results for three different networks: Shallow, AlexNet and VGG-16. We obtain the best results with the VGG-16 network, which is also the deepest network. This indicates learning from ranking makes it possible to train very deep networks efficiently without overfitting. These results are obtained by training from scratch, however we found that initializing the weights with a network pre-trained on ImageNet further improved the results. In the remainder of the experiments we thus use the VGG-16 network initialized with a pre-trained weights to train the ranking network in the following experiments.

\minisection{Baseline performance analysis.} In this experiment, we evaluate the effectiveness of using rankings to estimate image quality. We compare tree methods: fine-tuning the VGG-16 network initialized from ImageNet to obtain the mapping from images to their predicted scores (called Baseline), our method to train VGG-16 (initialized from ImageNet) on ranking database using all ranking dataset we generate (called RankIQA), and finally our RankIQA approach fine-tuned on the TID2013 database after training using ranked pairs of images (called RankIQA+FT). 

We follow the experimental protocol used in HOSA~\cite{xu2016blind}. The entire TID2013 database including all types of distortions is divided into 80\% training images and 20\% testing images according to the reference images. Thus, the same image can never appear in both training and test sets.  The results are shown in Table~\ref{ind_dis}, where ALL means testing all distortions together. All the experiments are performed 10 times and the average SROCC is reported\footnote{LCC results are provided in supplementary material.}.
From Table~\ref{ind_dis}, we can draw several conclusions. First, it is hard to obtain good results by training a deep network directly on IQA data. This is seen in the Baseline results and is due to the scarcity of training data. Second, our RankIQA method achieves superior results on almost all individual distortions even without ever using the TID2013 dataset -- which strongly demonstrates the effectiveness of training on ranking data. Slightly better results are obtained on ALL without comparing among different distortions during training the ranking network. The RankIQA-trained network alone does not provide accurate IQA scores (since it has never seen any) but does provide high correlation with the IQA scores as measured by SROCC.  After fine-tuning on the TID2013 database (RankIQA+FT), we considerably improve the ALL score, and improve the baseline by 16\%. However, in the fine-tuning process to optimize the ALL score the network balances the various distortions, and results decrease for several distortions. 

\subsection{Comparison with the state-of-the-art}

We compare the performance of our method using the VGG-16
network with state-of-the-art methods.  We perform experiments on the TID2013 and LIVE dataset.\footnote{Results on CSIQ~\cite{larson2010most} and MLIVE~\cite{jayaraman2012objective} are in supplementary material.}

\begin{table}
\begin{center}
\resizebox{\columnwidth}{!}{%
\begin{tabular}{|c|c|c|c|c|c|c|c|}
\hline
 & \textbf{LCC} & \textbf{JP2K}  & \textbf{JPEG}  & \textbf{GN} & \textbf{GB} & \textbf{FF} & \textbf{ALL} \\ \hline
\parbox[t]{2mm}{\multirow{4}{*}{\rotatebox[origin=c]{90}{\textbf{FR-IQA}}}}
& PSNR                             & 0.873 & 0.876 & 0.926 & 0.779 & 0.87  & 0.856 \\
& SSIM~\cite{wang2004image}        & 0.921 & 0.955 & 0.982 & 0.893 & 0.939 & 0.906 \\
& FSIM~\cite{zhang2011fsim}        & 0.91  & 0.985 & 0.976 & 0.978 & 0.912 & 0.96  \\
& DCNN~\cite{liang2016image}       & --  & -- & -- & -- & -- & 0.977  \\ \hline
\parbox[t]{2mm}{\multirow{8}{*}{\rotatebox[origin=c]{90}{\textbf{NR-IQA}}}}
& DIVINE~\cite{moorthy2011blind}   & 0.922 & 0.921 & 0.988 & 0.923 & 0.888 & 0.917 \\
& BLIINDS-II~\cite{saad2012blind}  & 0.935 & 0.968 & 0.98  & 0.938 & 0.896 & 0.93  \\
& BRISQUE~\cite{mittal2012no}      & 0.923 & 0.973 & 0.985 & 0.951 & 0.903 & 0.942 \\
& CORNIA~\cite{ye2012unsupervised} & 0.951 & 0.965 & 0.987 & 0.968 & 0.917 & 0.935 \\
& CNN~\cite{kang2014convolutional} & 0.953 & 0.981 & 0.984 & 0.953 & 0.933 & 0.953 \\
& SOM~\cite{zhang2015som}          & 0.952 & 0.961 & 0.991 & 0.974 & 0.954 & 0.962  \\
& DNN~\cite{bosse2016deep}          & -- & -- & -- & -- & -- & 0.972 \\ \cline{2-8}
& \textbf{RankIQA+FT}  & \textbf{0.975} & \textbf{0.986} & \textbf{0.994} & \textbf{0.988} & \textbf{0.960} & \textbf{0.982}  \\
\hline
\hline
\hline
& \textbf{SROCC} & \textbf{JP2K}  & \textbf{JPEG}  & \textbf{GN} & \textbf{BLUR}  & \textbf{FF}    & \textbf{ALL}   \\ \hline
\parbox[t]{2mm}{\multirow{4}{*}{\rotatebox[origin=c]{90}{\textbf{FR-IQA}}}}
& PSNR                             & 0.87  & 0.885 & 0.942 & 0.763 & 0.874 & 0.866 \\
& SSIM~\cite{wang2004image}        & 0.939 & 0.946 & 0.964 & 0.907 & 0.941 & 0.913 \\
& FSIM~\cite{liang2016image}       & 0.97  & 0.981 & 0.967 & 0.972 & 0.949 & 0.964 \\
& DCNN~\cite{liang2016image}       & --  & -- & -- & -- & -- & 0.975 \\ \hline
\parbox[t]{2mm}{\multirow{8}{*}{\rotatebox[origin=c]{90}{\textbf{NR-IQA}}}}
& DIVINE~\cite{moorthy2011blind}   & 0.913 & 0.91  & 0.984 & 0.921 & 0.863 & 0.916 \\
& BLIINDS-II~\cite{saad2012blind}  & 0.929 & 0.942 & 0.969 & 0.923 & 0.889 & 0.931 \\
& BRISQUE~\cite{mittal2012no}      & 0.914 & 0.965 & 0.979 & 0.951 & 0.887 & 0.94  \\
& CORNIA~\cite{ye2012unsupervised} & 0.943 & 0.955 & 0.976 & 0.969 & 0.906 & 0.942 \\
& CNN~\cite{kang2014convolutional} & 0.952 & 0.977 & 0.978 & 0.962 & 0.908 & 0.956 \\
& SOM~\cite{zhang2015som}          & 0.947 & 0.952 & 0.984 & 0.976 & 0.937 & 0.964 \\
&DNN~\cite{bosse2016deep}          & -- & -- & -- & -- & -- & 0.960 \\ \cline{2-8}
& \textbf{RankIQA+FT}  & \textbf{0.970} & \textbf{0.978} & \textbf{0.991} & \textbf{0.988} &  \textbf{0.954} & \textbf{0.981} \\ \hline
\end{tabular}}
\end{center}
\caption{LCC (above) and SROCC (below) evaluation on the LIVE
  dataset. We divide approaches into Full-reference (FR-IQA) and
  No-reference (IQA) techniques.}
\label{lcc}
\end{table}

\minisection{Evaluation on TID2013.} Table~\ref{ind_dis} also includes results of state-of-the-art methods. We see that for several very challenging distortions (14 to 18), where all other methods fail, we obtain satisfactory results. For individual distortions, there is a huge gap between our RankIQA method and other IQA methods on most distortions. The state-of-the-art method HOSA performs slightly better than our methods on 6 out of 24 distortions. For all distortions, our method RankIQA+FT achieves about 5\% higher than HOSA. Our methods perform well on distortions which are not included when training the ranking network, which indicates that different distortions share some common representation and training the network jointly on all distortions.

\minisection{Evaluation on LIVE.}  As done
in~\cite{kang2014convolutional,zhang2015som}, we randomly split the reference images on LIVE dataset
into 80\% training samples and 20\% testing, and compute the average
LCC and SROCC scores on the testing set after training to
convergence. This process is repeated ten times and the results are
averaged. These results are shown in Table~\ref{lcc}. The best method
for each dataset is indicated in bold. 
The column indicated with ALL means we combine all
five distortions together on LIVE dataset to train and test the
model. For fair comparison with the state-of-the-art, we train our
ranking model on four distortions except FF, but we fine-tune our
model on all five distortions in the LIVE dataset to compute ALL. Our approach
achieves about 1\% better than the best results reported on ALL
distortions for LCC. Similar conclusions are obtained for SROCC. This indicates that our method outperforms existing work
including the current state-of-the-art NR-IQA method SOM~\cite{zhang2015som} and DNN~\cite{bosse2016deep}, and also state-of-the-art FR-IQA method DCNN~\cite{liang2016image}. \emph{To the best of our knowledge this is the
  first time that an NR-IQA method surpasses the performance of FR-IQA
  methods (which have access to the undistorted reference image) on all LIVE distortions using the LCC and SROCC evaluation method}.

\subsection{Independence from IQA training data}

This final experiment is to demonstrate that our framework can be also
trained on non-IQA datasets. In the previous experiment the network is
trained from high-quality images of the Waterloo dataset. Instead here
we use the validation set of the Places2 dataset to generate ranked
images in place of the Waterloo dataset. The Places2 dataset is of
lower quality than Waterloo and is not designed for IQA research.  As
in the previous experiment, the final image quality scores are
predicted by fine-tuning on the LIVE dataset.  The performance of this
model is compared with the results trained on Waterloo in
Table~\ref{obj6}. The SROCC and LCC values are very similar,
demonstrating that our approach can be learnt from arbitrary, non-IQA
data.


\begin{table}
\begin{center}

\resizebox{0.9\columnwidth}{!}{%
\begin{tabular}{|c|c|c|c|c|c|}
\hline
\textbf{LCC} & \textbf{JP2K} & \textbf{JPEG}  & \textbf{GN} & \textbf{GB} & \textbf{ALL}  \\ \hline
RankIQA+FT (Waterloo)    & 0.975   & 0.986 & 0.994 & 0.988 & 0.982   \\
RankIQA+FT (Places2)    & 0.983   & 0.983 & 0.993  & 0.990 & 0.981 \\ \hline

\hline \hline
\textbf{SROCC} & \textbf{JP2K} & \textbf{JPEG}  & \textbf{GN} & \textbf{GB} & \textbf{ALL}  \\ \hline
RankIQA+FT (Waterloo)    & 0.970 & 0.978 & 0.991 & 0.988  & 0.981   \\
RankIQA+FT (Places2)    & 0.970   & 0.982 & 0.990  & 0.988 & 0.980 \\ \hline
\end{tabular}
}

\end{center}
\caption{SROCC and LCC results of models trained on the Waterloo
  and Places2 datasets, testing on LIVE.}
\label{obj6}
\end{table}

\section{Conclusions}
To address the scarcity of IQA data we have proposed a 
method which learns from ranked image datasets. Since this data can be generated in abundance we can
train deeper and wider networks than previous work. In addition, we have proposed a method for
efficient backpropagation in Siamese networks which circumvents the
need for hard-negative mining. Comparison with standard
pair sampling and hard-negative sampling shows
that our method converges faster and to a lower loss.  Results on LIVE and TID2013 datasets
 show that our NR-IQA approach
obtains superior results compared to existing NR-IQA techniques and
even FR-IQA methods. 

\minisection{Acknowledgements}
We acknowledge  the  Spanish project TIN2016-79717-R, the CHISTERA project M2CR (PCIN-2015-251) and the CERCA Programme / Generalitat de Catalunya. Xialei Liu acknowledges the Chinese Scholarship Council (CSC) grant No.201506290018. We also acknowledge the generous GPU donation from NVIDIA.


{\small
\bibliographystyle{ieee}
\bibliography{egbib}

\begin{thebibliography}{10}\itemsep=-1pt

\bibitem{bianco2016use}
S.~Bianco, L.~Celona, P.~Napoletano, and R.~Schettini.
\newblock On the use of deep learning for blind image quality assessment.
\newblock {\em arXiv preprint arXiv:1602.05531}, 2016.

\bibitem{bosse2016deep}
S.~Bosse, D.~Maniry, T.~Wiegand, and W.~Samek.
\newblock A deep neural network for image quality assessment.
\newblock In {\em Image Processing (ICIP), 2016 IEEE International Conference
  on}, pages 3773--3777. IEEE, 2016.

\bibitem{chen2009ranking}
W.~Chen, T.-Y. Liu, Y.~Lan, Z.-M. Ma, and H.~Li.
\newblock Ranking measures and loss functions in learning to rank.
\newblock In {\em Advances in Neural Information Processing Systems}, pages
  315--323, 2009.

\bibitem{chetouani2010novel}
A.~Chetouani, A.~Beghdadi, S.~Chen, and G.~Mostafaoui.
\newblock A novel free reference image quality metric using neural network
  approach.
\newblock In {\em Proc. Int. Workshop Video Process. Qual. Metrics Cons.
  Electrn}, pages 1--4, 2010.

\bibitem{chopra2005learning}
S.~Chopra, R.~Hadsell, and Y.~LeCun.
\newblock Learning a similarity metric discriminatively, with application to
  face verification.
\newblock In {\em Computer Vision and Pattern Recognition, 2005. CVPR 2005.
  IEEE Computer Society Conference on}, volume~1, pages 539--546. IEEE, 2005.

\bibitem{gao2015learning}
F.~Gao, D.~Tao, X.~Gao, and X.~Li.
\newblock Learning to rank for blind image quality assessment.
\newblock {\em Neural Networks and Learning Systems, IEEE Transactions on},
  26(10):2275--2290, 2015.

\bibitem{golestaneh2014no}
S.~A. Golestaneh and D.~M. Chandler.
\newblock No-reference quality assessment of jpeg images via a quality
  relevance map.
\newblock {\em Signal Processing Letters, IEEE}, 21(2):155--158, 2014.

\bibitem{he2015deep}
K.~He, X.~Zhang, S.~Ren, and J.~Sun.
\newblock Deep residual learning for image recognition.
\newblock In {\em Proceedings of the IEEE Conference on Computer Vision and
  Pattern Recognition}, 2016.

\bibitem{jayaraman2012objective}
D.~Jayaraman, A.~Mittal, A.~K. Moorthy, and A.~C. Bovik.
\newblock Objective quality assessment of multiply distorted images.
\newblock In {\em Signals, Systems and Computers (ASILOMAR), 2012 Conference
  Record of the Forty Sixth Asilomar Conference on}, pages 1693--1697. IEEE,
  2012.

\bibitem{jia2014caffe}
Y.~Jia, E.~Shelhamer, J.~Donahue, S.~Karayev, J.~Long, R.~Girshick,
  S.~Guadarrama, and T.~Darrell.
\newblock Caffe: Convolutional architecture for fast feature embedding.
\newblock {\em arXiv preprint arXiv:1408.5093}, 2014.

\bibitem{kang2014convolutional}
L.~Kang, P.~Ye, Y.~Li, and D.~Doermann.
\newblock Convolutional neural networks for no-reference image quality
  assessment.
\newblock In {\em Proceedings of the IEEE Conference on Computer Vision and
  Pattern Recognition}, pages 1733--1740, 2014.

\bibitem{kang2015simultaneous}
L.~Kang, P.~Ye, Y.~Li, and D.~Doermann.
\newblock Simultaneous estimation of image quality and distortion via
  multi-task convolutional neural networks.
\newblock In {\em Image Processing (ICIP), 2015 IEEE International Conference
  on}, pages 2791--2795. IEEE, 2015.

\bibitem{katsaggelos2012digital}
A.~K. Katsaggelos.
\newblock {\em Digital image restoration}.
\newblock Springer Publishing Company, Incorporated, 2012.

\bibitem{krizhevsky2012imagenet}
A.~Krizhevsky, I.~Sutskever, and G.~E. Hinton.
\newblock Imagenet classification with deep convolutional neural networks.
\newblock In {\em Advances in neural information processing systems}, pages
  1097--1105, 2012.

\bibitem{larson2010most}
E.~C. Larson and D.~M. Chandler.
\newblock Most apparent distortion: full-reference image quality assessment and
  the role of strategy.
\newblock {\em Journal of Electronic Imaging}, 19(1):011006--011006, 2010.

\bibitem{lecun1998gradient}
Y.~LeCun, L.~Bottou, Y.~Bengio, and P.~Haffner.
\newblock Gradient-based learning applied to document recognition.
\newblock {\em Proceedings of the IEEE}, 86(11):2278--2324, 1998.

\bibitem{liang2016image}
Y.~Liang, J.~Wang, X.~Wan, Y.~Gong, and N.~Zheng.
\newblock Image quality assessment using similar scene as reference.
\newblock In {\em European Conference on Computer Vision}, pages 3--18.
  Springer, 2016.

\bibitem{liu2014no}
L.~Liu, H.~Dong, H.~Huang, and A.~C. Bovik.
\newblock No-reference image quality assessment in curvelet domain.
\newblock {\em Signal Processing: Image Communication}, 29(4):494--505, 2014.

\bibitem{mittal2012no}
A.~Mittal, A.~K. Moorthy, and A.~C. Bovik.
\newblock No-reference image quality assessment in the spatial domain.
\newblock {\em Image Processing, IEEE Transactions on}, 21(12):4695--4708,
  2012.

\bibitem{moorthy2010two}
A.~K. Moorthy and A.~C. Bovik.
\newblock A two-step framework for constructing blind image quality indices.
\newblock {\em Signal Processing Letters, IEEE}, 17(5):513--516, 2010.

\bibitem{moorthy2011blind}
A.~K. Moorthy and A.~C. Bovik.
\newblock Blind image quality assessment: From natural scene statistics to
  perceptual quality.
\newblock {\em IEEE Transactions on Image Processing}, 20(12):3350--3364, 2011.

\bibitem{ponomarenko2013color}
N.~Ponomarenko, O.~Ieremeiev, V.~Lukin, K.~Egiazarian, L.~Jin, J.~Astola,
  B.~Vozel, K.~Chehdi, M.~Carli, F.~Battisti, et~al.
\newblock Color image database tid2013: Peculiarities and preliminary results.
\newblock In {\em Visual Information Processing (EUVIP), 2013 4th European
  Workshop on}, pages 106--111. IEEE, 2013.

\bibitem{saad2012blind}
M.~A. Saad, A.~C. Bovik, and C.~Charrier.
\newblock Blind image quality assessment: A natural scene statistics approach
  in the dct domain.
\newblock {\em Image Processing, IEEE Transactions on}, 21(8):3339--3352, 2012.

\bibitem{schroff2015facenet}
F.~Schroff, D.~Kalenichenko, and J.~Philbin.
\newblock Facenet: A unified embedding for face recognition and clustering.
\newblock In {\em Proceedings of the IEEE Conference on Computer Vision and
  Pattern Recognition}, pages 815--823, 2015.

\bibitem{sculley2009large}
D.~Sculley.
\newblock Large scale learning to rank.
\newblock In {\em NIPS Workshop on Advances in Ranking}, pages 1--6, 2009.

\bibitem{sharifi1995estimation}
K.~Sharifi and A.~Leon-Garcia.
\newblock Estimation of shape parameter for generalized gaussian distributions
  in subband decompositions of video.
\newblock {\em Circuits and Systems for Video Technology, IEEE Transactions
  on}, 5(1):52--56, 1995.

\bibitem{sheikh2006statistical}
H.~R. Sheikh, M.~F. Sabir, and A.~C. Bovik.
\newblock A statistical evaluation of recent full reference image quality
  assessment algorithms.
\newblock {\em Image Processing, IEEE Transactions on}, 15(11):3440--3451,
  2006.

\bibitem{live2}
H.~R. Sheikh, Z.~Wang, L.~Cormack, and A.~C. Bovik.
\newblock Live image quality assessment database.
\newblock \url{http://live.ece.utexas.edu/research/quality}.

\bibitem{simo2015discriminative}
E.~Simo-Serra, E.~Trulls, L.~Ferraz, I.~Kokkinos, P.~Fua, and F.~Moreno-Noguer.
\newblock Discriminative learning of deep convolutional feature point
  descriptors.
\newblock In {\em Proceedings of the IEEE International Conference on Computer
  Vision}, pages 118--126, 2015.

\bibitem{simonyan2014very}
K.~Simonyan and A.~Zisserman.
\newblock Very deep convolutional networks for large-scale image recognition.
\newblock {\em ICLR}, 2015.

\bibitem{song2015deep}
H.~O. Song, Y.~Xiang, S.~Jegelka, and S.~Savarese.
\newblock Deep metric learning via lifted structured feature embedding.
\newblock In {\em Proceedings of the IEEE Conference on Computer Vision and
  Pattern Recognition}, 2016.

\bibitem{van2006image}
J.~Van~Ouwerkerk.
\newblock Image super-resolution survey.
\newblock {\em Image and Vision Computing}, 24(10):1039--1052, 2006.

\bibitem{wang2015unsupervised}
X.~Wang and A.~Gupta.
\newblock Unsupervised learning of visual representations using videos.
\newblock In {\em Proceedings of the IEEE International Conference on Computer
  Vision}, pages 2794--2802, 2015.

\bibitem{wang2002image}
Z.~Wang, A.~C. Bovik, and L.~Lu.
\newblock Why is image quality assessment so difficult?
\newblock In {\em Acoustics, Speech, and Signal Processing (ICASSP), 2002 IEEE
  International Conference on}, volume~4, pages IV--3313. IEEE, 2002.

\bibitem{wang2004image}
Z.~Wang, A.~C. Bovik, H.~R. Sheikh, and E.~P. Simoncelli.
\newblock Image quality assessment: from error visibility to structural
  similarity.
\newblock {\em IEEE transactions on image processing}, 13(4):600--612, 2004.

\bibitem{xu2016blind}
J.~Xu, P.~Ye, Q.~Li, H.~Du, Y.~Liu, and D.~Doermann.
\newblock Blind image quality assessment based on high order statistics
  aggregation.
\newblock {\em IEEE Transactions on Image Processing}, 25(9):4444--4457, 2016.

\bibitem{yan2014learning}
J.~Yan, S.~Lin, S.~B. Kang, and X.~Tang.
\newblock A learning-to-rank approach for image color enhancement.
\newblock In {\em Computer Vision and Pattern Recognition (CVPR), 2014 IEEE
  Conference on}, pages 2987--2994. IEEE, 2014.

\bibitem{yan2013no}
Q.~Yan, Y.~Xu, and X.~Yang.
\newblock No-reference image blur assessment based on gradient profile
  sharpness.
\newblock In {\em Broadband Multimedia Systems and Broadcasting (BMSB), 2013
  IEEE International Symposium on}, pages 1--4. IEEE, 2013.

\bibitem{ye2012no}
P.~Ye and D.~Doermann.
\newblock No-reference image quality assessment using visual codebooks.
\newblock {\em Image Processing, IEEE Transactions on}, 21(7):3129--3138, 2012.

\bibitem{ye2012unsupervised}
P.~Ye, J.~Kumar, L.~Kang, and D.~Doermann.
\newblock Unsupervised feature learning framework for no-reference image
  quality assessment.
\newblock In {\em Computer Vision and Pattern Recognition (CVPR), 2012 IEEE
  Conference on}, pages 1098--1105. IEEE, 2012.

\bibitem{zagoruyko2015learning}
S.~Zagoruyko and N.~Komodakis.
\newblock Learning to compare image patches via convolutional neural networks.
\newblock In {\em Proceedings of the IEEE Conference on Computer Vision and
  Pattern Recognition}, pages 4353--4361, 2015.

\bibitem{zhang2011fsim}
L.~Zhang, L.~Zhang, X.~Mou, and D.~Zhang.
\newblock Fsim: a feature similarity index for image quality assessment.
\newblock {\em IEEE transactions on Image Processing}, 20(8):2378--2386, 2011.

\bibitem{zhang2015som}
P.~Zhang, W.~Zhou, L.~Wu, and H.~Li.
\newblock Som: Semantic obviousness metric for image quality assessment.
\newblock In {\em Proceedings of the IEEE Conference on Computer Vision and
  Pattern Recognition}, pages 2394--2402, 2015.

\bibitem{zhang2015label}
Q.~Zhang, Z.~Ji, S.~Lilong, and I.~Ovsiannikov.
\newblock Label-free non-reference image quality assessment via deep neural
  network, Nov.~3 2015.
\newblock US Patent App. 14/931,843.

\bibitem{places}
B.~Zhou, A.~Lapedriza, J.~Xiao, A.~Torralba, and A.~Oliva.
\newblock Learning deep features for scene recognition using places database.
\newblock In {\em Advances in neural information processing systems}, pages
  487--495, 2014.

\end{thebibliography}
}

\end{document}